\documentclass[%
 aip,apl,
 reprint,
]{revtex4-1}

\usepackage{graphicx}% Include figure files
\usepackage{dcolumn}% Align table columns on decimal point
\usepackage{bm}% bold math
\usepackage{float}
\usepackage[utf8]{inputenc}
\usepackage[T1]{fontenc}
\usepackage{mathptmx}
\usepackage{booktabs}
\begin{document}

\preprint{AIP/123-QED}

\title[Scanning Probe State Recognition With Multi-Class Neural Network Ensembles]{Scanning Probe State Recognition\\ With Multi-Class Neural Network Ensembles}
% Force line breaks with \\

\author{O. Gordon}
 	\affiliation{School of Physics \& Astronomy, The University of Nottingham, University Park, Nottingham, NG7 2RD, United Kingdom.}
	%\email{oliver.gordon@nottingham.ac.uk}
\author{P. D'Hondt}%
 	\affiliation{School of Physics \& Astronomy, The University of Nottingham, University Park, Nottingham, NG7 2RD, United Kingdom.}
 	\affiliation{IEMN – Laboratoire Central De L'Institut, Cité Scientifique, Avenue Henri Poincaré, CS 60069, 59 652 Villeneuve d'Ascq Cedex, France.}
\author{L. Knijff}
\affiliation{Debye Institute for Nanomaterials Science, Utrecht University, Utrecht 3584 CC, Netherlands.}
\author{S. Freeney}
\affiliation{Debye Institute for Nanomaterials Science, Utrecht University, Utrecht 3584 CC, Netherlands.}
\author{F. Junqueira}
	\affiliation{School of Physics \& Astronomy, The University of Nottingham, University Park, Nottingham, NG7 2RD, United Kingdom.}
	\email{Filipe.Junqueira@nottingham.ac.uk}
\author{P. Moriarty}
\affiliation{School of Physics \& Astronomy, The University of Nottingham, University Park, Nottingham, NG7 2RD, United Kingdom.}
\email{philip.moriarty@nottingham.ac.uk}
\author{I. Swart}
	\affiliation{Debye Institute for Nanomaterials Science, Utrecht University, Utrecht 3584 CC, Netherlands.}

\date{\today}% It is always \today, today,

\begin{abstract}
One of the largest obstacles facing scanning probe microscopy is the constant need to correct flaws in the scanning probe \textit{in situ}. This is currently a manual, time-consuming process that would benefit greatly from automation. Here we introduce a convolutional neural network protocol that enables automated recognition of a variety of desirable and undesirable scanning probe tip states on both metal and non-metal surfaces. By combining the best performing models into majority voting ensembles, we find that the desirable states of H:Si(100) can be distinguished with a mean precision of 0.89 and an average receiver-operator-characteristic curve area of 0.95. More generally, high and low-quality tips can be distinguished with a mean precision of 0.96 and near perfect area-under-curve of 0.98. With trivial modifications, we also successfully automatically identify undesirable, non-surface-specific states on surfaces of Au(111) and Cu(111). In these cases we find mean precisions of 0.95 and 0.75 and area-under-curves of 0.98 and 0.94, respectively.
\end{abstract}

\maketitle

\section{\label{sec:Intro}Introduction}

	Whilst scanning probe microscopy (SPM) has allowed researchers to make observations at the atomic level for decades\cite{binnig1982surface,Gewirth1997,Voigtlaender2016}, success is highly reliant on the production of atomically sharp scanning tips. Although sharp tips are readily created \textit{ex situ}\cite{Li2016,Rezeq2006}, imperfections in the tip apex including the presence of "double" or multiple tips mean that image artefacts often appear spontaneously during experimental sessions. To maintain resolution, apex flaws must be repeatedly corrected \textit{in situ} through a repeated combination of controlled voltage pulsing and/or tip crashing. 

	Despite the manual, time-consuming nature of tip correction, there have been surprisingly few attempts to date to automate the process\cite{Straton2014,Wang2016,Woolley2011,Rashidi2018}. Of these attempts, a variety of pitfalls have been identified, ranging from low accuracy and high computational cost to faltering when multiple tip flaws are present. They also often require a degree of manual input, are invariant to scale and rotation, or fail when the tip spontaneously changes the visible resolution mid-image. Convolutional neural networks (CNNs) are a highly promising candidate for this task, which routinely achieve high accuracy in complex vision tasks such as medical, satellite and digit recognition\cite{Ciresan2013,Lochner2016,Krizhevsky2012}. Despite this, in the context of SPM only Rashidi and Wolkow\cite{Rashidi2018} have to the best of our knowledge used CNNs for tip-conditioning, and only while scanning the H:Si(100) surface. 
	
	In this paper, we broaden Rashidi and Wolkow's approach to a method that can reliably assess the state of an SPM tip while scanning on both metallic and semiconducting surfaces. This is achieved via majority voting from an ensemble of multiple CNNs.	We also increase the number of distinct recognisable states and allow for desirable and non-desirable tip state classifications at non-fixed length scales and rotations. We ultimately present ensemble CNNs capable of classifying multiple tip states with human-like performance and weighted accuracies in excess of 80\% (and 90\% in some cases).

\section{\label{sec:Methods}Methods}

\begin{figure*}
		\centering
		\includegraphics[width=\textwidth]{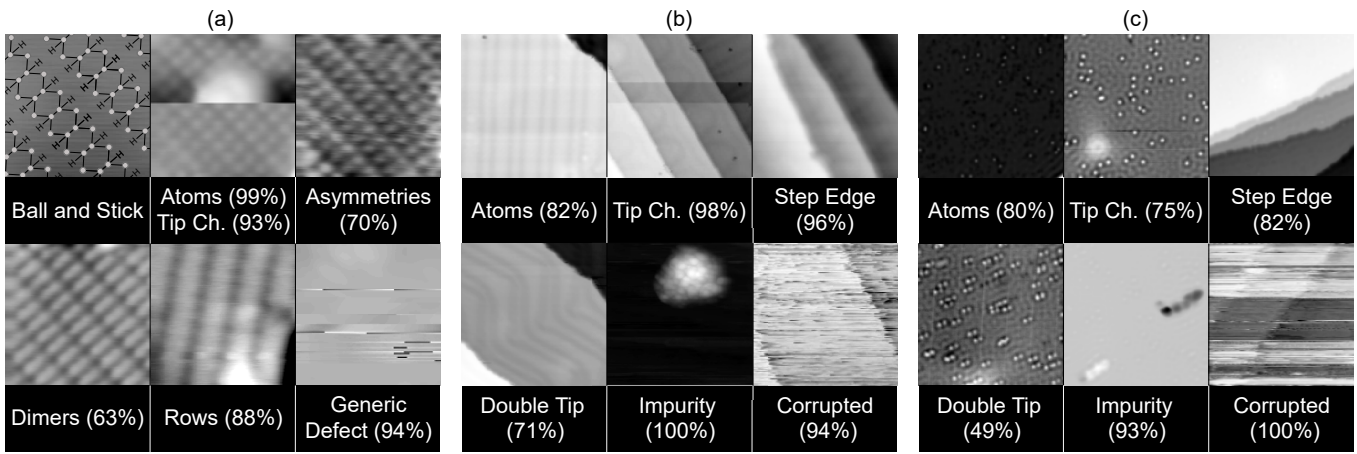}
		\label{fig:all_res:atoms}
		\caption{Selection of images demonstrating key tip states for STM imaging of (a) H:Si(100), (b) Au(111), (c) Cu(111), and the confidence thresholds of convolutional neural networks used to classify them. We note that in many examples, features can appear to strongly blend between images, such as with asymmetries and dimer-like modulation in rows in (a). Because creating unambiguous training data was impractical, we therefore combined these classes. }\label{fig:all_res}
	\end{figure*}

	When assessing the quality of an SPM image, not all features are considered equally. An operator may want to observe desirable features, but actively avoid undesirable artefacts. While Rashidi and Wolkow\cite{Rashidi2018} distinguished two key tip states when imaging H:Si(100) for CNN-driven automated SPM, a much wider set of classifications is possible. For this surface these include\cite{Sweetman2012}: 'atoms' (for the sharpest tips), 'dimers', 'asymmetries', and 'rows'. Example images are shown in Figure \ref{fig:all_res}. Although H:Si(100) is a substrate that underpins many advances in single atom technologies\cite{Moller2017,Fuechsle2012,Lopinski2000}, these classifications are surface-specific. 'Double tips', 'tip changes', 'step edges', 'impurities' and image corruption 'defects' are all undesirable artefacts that could apply to any surface. To this end, and to demonstrate the general applicability of our CNN protocol, we also study two other commonly studied surfaces\cite{Sun2011,Jacobse2016}: Cu(111) and Au(111).

	To train the CNNs, 13789 images were first obtained. H:Si(100) images were acquired at room temperature between March 2014 and November 2015 on a Scienta Omicron Variable Temperature STM at various rotations, length scales between 3x3nm\textsuperscript{2} and 80x80nm\textsuperscript{2}, and resolutions up to 512x512 pixels. The Au(111) and Cu(111) images were acquired similarly on an Omicron LT, but at a fixed scan size (30x30nm), resolution (150x150 pixels), and temperature (4.5K). Minimal processing\cite{Stirling2013} was applied to bring most data on the order -1 to 1. The images of H:Si(100) were manually classified into the four tip states listed above (i.e. atoms, dimers, asymmetries, and rows) and two other categories: tip changes and generic defects. Similarly, the Au(111) and Cu(111) images were classified into five categories of undesirable defects, and the one desirable state of sharp resolution\footnote{The Au(111) and Cu(111) data were acquired by the Utrecht group, whereas the H:Si(100) images were obtained at Nottingham and classified separately and by different researchers.}. To prevent overestimation of the final accuracies, a random selection of images were withheld as holdout data for analysis. Train and test data were then created by randomly splitting the remaining images with a 80/20 ratio. 
	
	Although in practice SPM images are multi-label (in which images can belong to multiple categories), we classified and discarded data such that we only trained with multi-class (in which images can belong to only one category). This was beneficial as CNNs learn from the relationship between categories, so did not have to learn to ignore relations that did not exist. It is also known that although a CNN can learn with ambiguous or misleading training labels, performance is reduced\cite{Jindal2016, Frenay2014, Zhu2004}. However, because undesirable tip changes can occur even when observing a desirable tip state, these were not excluded. Instead, tip changes were trained in a separate binary yes/no CNN for H:Si(100), and the remaining Si images trained in a four-class CNN. Tip change separation was not applied to the Cu and Au datasets, as the aim was to explore the relations between undesirable defects.
	
	There was also a great deal of variety between classifications despite the consistent classification scheme and limited number of classifiers. Whilst we did not train on images that the human classifiers did not agree on, the large degree of ambiguity in classification meant that many ambiguous images remained. In the absence of a perfect classification system and greater number of classifiers, these imperfect human classifications formed the training data that the network had to learn from. As such, no CNN could achieve 100\% accuracy without overfitting. For example, given that 78\% of the silicon dataset was agreed upon, it could be tentatively argued that a human-like CNN would score similarly. (A poll carried out in our group which involved the manual classification of a small subset of the Si dataset by nine scanning probe microscopists, similarly found only 73\% agreement). Ultimately, 3386 H:Si(100), 3600 Cu(111) and 2470 Au(111) images were used for training/testing and 431, 1120 and 432 images for verification, respectively.
	
	To improve training performance, the training and testing data were repeated and augmented. Expanding on the simple vertical and horizontal flips used by Rashidi, we also applied rotations from 0-360$^\circ$, cropped, panned, and added random amounts of Gaussian noise. This improves performance by reducing overfitting, in which a CNN learns to classify training data at the expense of misclassifying unseen data\cite{Rasmussen2004}. For the tip change categories, only horizontal flips and Gaussian noise were applied as in our case tip changes were horizontal shears, and zooming in might crop off the discontinuity. 
	
	The CNNs also had to be prevented from overfitting by learning about the differing number of images in each class. For example, 5.6\% of the images in the H:Si(100) filtered dataset were atoms, compared to 41.9\% generic defects. Similar variety was also observed in the Cu(111) and Au(111) sets. We therefore weighted each category by the reciprocal of the percentage of each class present, used a weighted accuracy metric\cite{Pedregosa2011} (where the reciprocal of number of classes is defined as guessing), and randomly shuffled data. Without these steps, a CNN could rapidly take an example dataset containing nine good images and one bad, and be 90\% accurate by guessing all images as good. It is for this reason that other authors warn against using solely accuracy to judge the performance of weighted datasets\cite{Saito2015,Rashidi2018}.
	
	Furthermore, our priority when establishing a CNN protocol was not to maximise the ratio of true to false classifications (i.e. accuracy), but to maximise true \textit{positive} classifications. This was observed using the metric of precision (defined as the ratio of true positive classifications to total positive classifications). By increasing the confidence threshold required to make a positive classification, precision was increased at the cost of increased false negatives and therefore decreased accuracy. This is visible in receiver operator characteristic (ROC) and precision-recall (PR) curves. All-round performance is given by the area under ROC (AUROC), in which a perfect classifier has an AUROC of 1, and guessing 0.5. These metrics are also not affected by class imbalance\cite{Saito2015}, and were therefore superior to accuracy. (We therefore also made the traditional distinction between accuracy and precision, rather than used the terms interchangeably\cite{Rashidi2018}).
		
	In addition to the network described by Rashidi and Wolkow\cite{Rashidi2018} (RW), we tested models similar to the popular visual geometry group (VGG) network with and without batch normalisation. We also tested a model highly similar to Google's Squeezenet\cite{Iandola2017}. This network had ten back-to-back convolutional layers, filters increasing in number from 32 to 1024, strides alternating between 1 and 2, and 3x3 convolutions. Between layers, batch normalisation and the elu\cite{Clevert2015} activation function were applied. Loss rate was also gradually reduced during training to reduce overfitting further. 
	
	We note that although multi-class networks are typically trained with a sigmoid activation function and categorical cross-entropy loss function, we did not use these. Because future data would be multi-label, we instead opted for the multi-label standard of softmax and binary cross-entropy\cite{Duan2003}. This made the confidence prediction of each category 0-1 independent of each other, instead of mathematically linking all the predictions for each category to sum to 1. \footnote{We also note that although Rashidi and Wolkow used sigmoid and categorical cross-entropy functions\cite{Rashidi2018}, they were not the standard choices for their binary classification scheme because both the confidence of the positive and negative classification could be high, degrading performance.}  
	
	To determine an optimal ensemble model, a variety of model structures, optimisers\cite{opt_adam,opt_adadelta,opt_adagrad,opt_others} and learning rates, were trained and analysed for each model structure. In all cases, training was done at a batch size of 128, and image sizes of 128x128 pixels. At higher sizes training time massively increased, but with little to no improvement in performance. A more traditional Random Forest Classifier (RFC) was also implemented for comparison. The top performing models were then combined to create a majority voting ensemble, which have been shown to further improve performance\cite{Dietterich2000}. For ambiguous data, this was also more analogous to a majority human vote with different models having different preferences. 
	
\section{\label{sec:Results}Results}

	First, the individual models were compared. Table \ref{tab:models} displays the best results obtained for all the desirable/undesirable multi-class models. Although all networks performed significantly better than RFC and weighted random guessing, the RW CNN performed poorly and similar to the more traditional RFC. Furthermore, at the 32x32 image size described by Rashidi and Wolkow\cite{Rashidi2018}, RW performed comparable to random guessing, indicating the high difficulty of this task. 
	
	We also found a wide variety in performance between different surfaces,  indicating that \textit{CNN architectures respond differently to different surfaces}. For example, whilst Squeezenet was the best performer for H:Si(100), only VGG like networks were suitable for Au and Cu. Furthermore, batch normalisation improved performance on H:Si(100), whilst negatively impacting Au(111) and Cu(111). This variance is understandable, given the current lack of consensus on how network structure relates to performance on a given data set\cite{Szegedy2015}. 
	
\begin{table*}[!hbtp]\centering
	\caption{Table to compare the performance of a variety of machine learning methods for classifying desirable and undesirable tip states for six classes of Au(111) and Cu(111), and four classes of H:Si(100). The SqueezeNet, VGG, Rashidi-Wolkow (RW) and ensemble networks are examples of convolutional neural networks. These all performed significantly better than the more traditional Random Forest Classifier (RFC) with 5000 trees, and random guessing, which performed as expected.}
	\label{tab:models}
	\begin{tabular}{@{}lccccccccccccccccccccccccccc@{}}\toprule[2pt]
		& \multicolumn{3}{c}{\textbf{Ensemble}}
		& \phantom{a}& \multicolumn{3}{c}{\textbf{SqueezeNet}}
		& \phantom{a}& \multicolumn{3}{c}{\textbf{VGG (Batchnorm)}}
		& \phantom{a}& \multicolumn{3}{c}{\textbf{VGG}} 
		& \phantom{a}& \multicolumn{3}{c}{\textbf{RW}} 
		& \phantom{a}& \multicolumn{3}{c}{\textbf{RFC}}
		& \phantom{a}& \multicolumn{3}{c}{\textbf{Random}}\\
		\cmidrule{2-4} \cmidrule{6-8} \cmidrule{10-12} \cmidrule{14-16} \cmidrule{18-20} \cmidrule{22-24} \cmidrule{26-28}
		& Si & Au & Cu && Si & Au & Cu && Si & Au & Cu && Si & Au & Cu && Si & Au & Cu && Si & Au 
		& Cu && Si & Au & Cu\\ \midrule
		\textbf{{AUROC}} & 0.95 & 0.98 & 0.94 &&
						 0.94 & 0.95 & 0.88 &&
						 0.92 & 0.93 & 0.85 &&
						 0.91 & 0.98 & 0.93 &&
						 0.87 & 0.82 & 0.77 &&
						 0.79 & 0.88 & 0.83 &&
						 0.50 & 0.50 & 0.50\\
		\textbf{Bal. Acc.} & 0.78 & 0.86 & 0.80 &&
							0.77 & 0.71 & 0.67 &&
							0.74 & 0.74 & 0.59 &&
							0.72 & 0.86 & 0.76 &&
							0.62 & 0.55 & 0.50 &&
							0.46 & 0.53 & 0.52 &&
							0.25 & 0.16 & 0.16\\
		\textbf{{Precision}} & 0.89 & 0.95 & 0.75 &&
							0.88 & 0.82 & 0.67 &&
							0.82 & 0.77 & 0.57 &&
							0.82 & 0.92 & 0.72 &&
							0.71 & 0.54 & 0.47 &&
							0.57 & 0.62 & 0.51 &&
							0.25 & 0.18 & 0.17\\
		\bottomrule[2pt]
	\end{tabular}
\end{table*}

	From here, the best performing networks were taken and turned into an ensemble. Three were chosen as this gave a good balance between performance and memory usage. As expected, small performance improvements were seen when moving to ensembles. For H:Si(100), the top performer was an ensemble of two SqueezeNets and one batch-normalised VGG, with adam, sgd and rmsprop optimisers, and learning rates of 0.001, 0.0001 and 0.0001 respectively. However, our ensemble structure did not train well with Cu(111) and Au(111) (65\% balanced accuracy, 0.64 precision, 0.89 AUROC on Cu(111)). This is likely because of the low performance of the component networks on these surfaces. As such, ensembles for Au(111) and Cu(111) were therefore created from multiple repeats of the VGG like network. 
	
	Although Table \ref{tab:models} indicated strong overall performance, these numbers were likely underestimates. Considering Figure \ref{fig:all_res}, there was a high degree of feature overlap which made the classification task subjective. For example, some Si dimer images had a bright, asymmetric edge. While these categories were eventually combined as they were routinely misclassified together,\footnote{We note that although it seems like asymmetries/atoms should be grouped because they are visually similar, the end goal was to observe atoms.} the filtered multi-class images still contained multiple acceptable multi-label answers. Because the metrics only allowed for one classification for any given image, the network was often punished despite producing a sensible distribution. This would have been avoided were significantly more classifiers employed. 

\begin{figure}
		\includegraphics[width=0.47\textwidth]{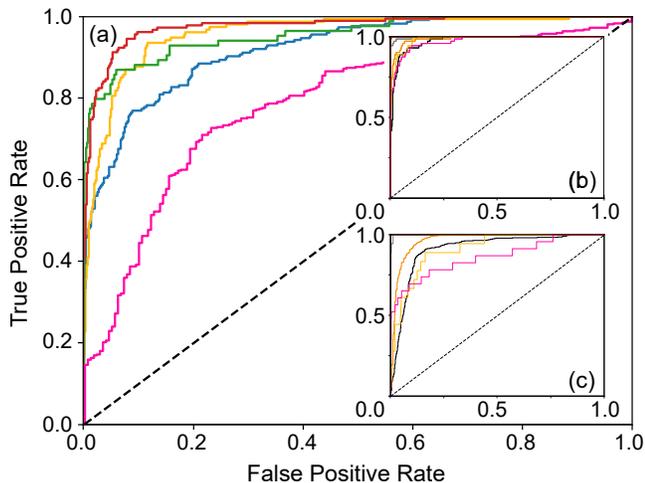}
		\caption{Receiver Operator Characteristic graphs demonstrating the overall performance and area under curve as the confidence threshold required to make a positive prediction is varied for CNN ensembles. Performance is compared for the classification of scanning probe images of (a) H:Si(100), (b) Au(111) and (c) Cu(111). A perfect classifier has an area under curve of 1, with guessing 0.50 (black dash, theoretical). For (a) we find asymmetry/dimer = 0.92 (blue), individual atoms = 0.96 (yellow), rows = 0.95 (green), tip change = 0.79 (pink), and generic defect = 0.98 (red). For (b) and (c) respectively we find impurities = 1.00, 1.00 (grey), double tip = 0.98, 0.91 (black), corruption = 1.00, 1.00 (brown), individual atoms = 0.98, 0.91 (yellow), step edges = 0.99, 0.97 (orange), and tip change = 0.97, 0.86 (pink).}
		\label{fig:ROC}
	\end{figure}

	In spite of this, Figure \ref{fig:ROC} shows that the AUROC for all categories and surfaces was very high. This indicated that the classifier had a low false positive rate, but at the cost of a high false negative rate. Although decreasing accuracy, this characteristic is not detrimental to areas such as ours when only positive predictions are to be acted upon. Furthermore, ambiguous classifications often had confidences $<$0.5, increasing false negative count and reducing accuracy further still. Unambiguous cases, such as corruptions, impurities and individual atoms of Au(111) and Cu(111) and generic defects of H:Si(100) were classified extremely well with near perfect AUROC.
	
\begin{figure}
		\includegraphics[width=0.47\textwidth]{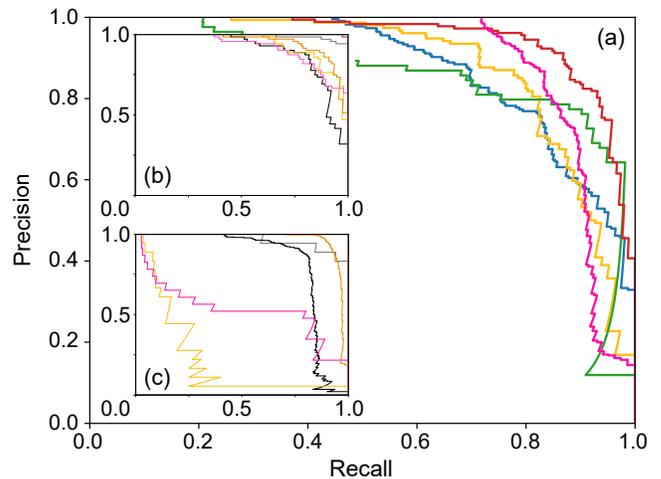}
		\caption{Precision-Recall graphs to demonstrate the overall performance of ensemble CNNs when classifying the known tip states for images of (a) H:Si(100) (b) Au(111) (c) Cu(111) as the confidence threshold required to make a positive classification was varied. Precision is the percentage of true positives compared to total positive classification, while recall is the percentage of positive classifications that have been correctly identified as positive. Some tip states are desirable and surface specific, such as asymmetry/dimer (blue), individual atoms (yellow), and rows (green). However, tip changes (pink), impurities (grey), double tips (black), corrupted (brown), step edges (orange), and generic defect (red), are undesirable. Performance is strong, except for individual atoms and tip change in (c).}
		\label{fig:PR}
	\end{figure}

	Furthermore, misclassifications were often between sets of desirable/undesirable states, rather than with desirable states being misclassified as undesirable and vice versa. To demonstrate this, the four class H:Si(100) and Au(111) ensembles were simplified into "good/bad" classifiers. They then achieved improved balanced accuracies of 93\% and 91\%, mean precisions of 0.96 and 0.97, and AUROCs of 0.98 and 0.98 respectively. Cu(111) did not improve owing to poor PR of individual atoms and tip changes, as visible in Figure \ref{fig:PR}(c).

	However, although tip changes were classified respectably with Au(111) and Cu(111), this was not the case with H:Si(100). When including the separate binary network to cover all classes for H:Si(100), performance was significantly poorer, with a balanced accuracy of 77\%, mean precision of 0.88, and average AUROC of 0.92. This is particularly visible in Figure \ref{fig:ROC}, with the tip change category having an ROC line below the other categories and AU of 0.80. This is likely because when augmentations were limited to simple flips and noise, the network rapidly overfit and learning had to be stopped earlier. Regardless, few false positives were made for tip changes when increasing confidence thresholds. This is because precision was only seen to decrease at high values of recall, as visible in Figure \ref{fig:PR}. As such, tip states could still be distinguished with a low false positive rate by requiring a high confidence threshold.

\section{Conclusion}
	
	We have successfully trained CNNs capable of classifying numerous desirable and undesirable SPM tip states for multiple surfaces. We achieve significantly greater all-round performance than other supervised learning techniques, and an even stronger ability to differentiate good and bad tip apices. However, we find that without significantly expanded datasets not all surfaces are equally suitable for CNN classification.
	
	Were more human classifiers available, the networks should have been trained on the entire multi-label dataset, and then scored based on a cross-entropy of average classifications. Performance could also be improved further with the addition of more training data, and potentially combined with time-dependent data to allow for real-time classification and tip enhancement during scanning. In its current state, the CNNs will enable a fully autonomous in situ approach to selecting and observing a variety of tip states during imaging, spectroscopic, and atomic manipulation experiments. 

\begin{acknowledgments}
	OG, FJ, PdH and PM acknowledge funding from the Engineering and Physical Sciences Research Council via grant EP/N02379X/1. IS acknowledges funding from NWO via grant 16PR3245. OG and PM also thank Bob Wolkow, Mohammad Rashidi and Jeremy Croshaw of the University of Alberta, and Ken Gordon, President and CEO of Quantum Silicon Inc. for sharing data and a series of very helpful personal communications.
\end{acknowledgments}

\nocite{*}
\bibliography{references}% Produces the bibliography via BibTeX.

\end{document}